\documentclass[a4paper,fleqn]{cas-sc}

\usepackage[american]{babel}
\usepackage{csquotes}
\usepackage{amsthm}
\usepackage[numbers,sort&compress]{natbib}
\usepackage{microtype}
\usepackage{xurl}
\usepackage{tabularx}
\usepackage[scaled=.85]{inconsolata}

\ExplSyntaxOn
\keys_set:nn { stm / mktitle } { nologo }
\ExplSyntaxOff
\RenewDocumentCommand\printorcid{}{}
\hypersetup{pdftitle={Bottleneck strata for bounds on probabilities of causation},pdfauthor={Marek Deja}}

\newtheorem{proposition}{Proposition}
\newtheorem{theorem}{Theorem}
\newtheorem{lemma}{Lemma}

\theoremstyle{definition}

\theoremstyle{remark}

\newcommand{\PNS}{\mathrm{PNS}}
\newcommand{\PN}{\mathrm{PN}}
\newcommand{\PS}{\mathrm{PS}}
\newcommand{\KL}{D}
\newcommand\mstablenote[1]{\par\vspace{3pt}{\footnotesize\leftskip=0pt\rightskip=0pt\parfillskip=0pt plus 1fil\noindent #1\par}}
\newcommand\msbacksection[1]{\section*{#1}}
\newcommand\msabstract[1]{#1}
\makeatletter
\g@addto@macro\appendix{\renewcommand\p@section{Appendix~}\renewcommand\p@subsection{Appendix~}}
\makeatother

\begin{document}
\let\WriteBookmarks\relax
\def\floatpagepagefraction{1}
\def\textpagefraction{.001}

\shorttitle{Bottleneck strata for bounds on probabilities of causation}
\shortauthors{M. Deja}

\title [mode = title]{Bottleneck strata for bounds on probabilities of causation from combined experimental and observational data}

\author[1]{Marek Deja}
\cormark[1]
\ead{marek.deja@uj.edu.pl}

\affiliation[1]{organization={Institute of Information Studies, Jagiellonian University},
  addressline={ul. prof. Stanisława Łojasiewicza 4},
  city={Kraków},
  postcode={30-348},
  country={Poland}}

\cortext[1]{Corresponding author}

\begin{abstract}
\msabstract{Probabilities of causation, such as the probability that a treatment was necessary and sufficient for an outcome, are partially identified when experimental and observational data are combined, and pre-treatment covariates narrow the bounds when every stratum holds enough units of every design arm. The method learns a few strata with the information bottleneck, whose relevance variable is the outcome augmented with the design arm and the received treatment in a two-stage randomized preference design. Every stratifier that depends on pre-treatment covariates gives valid bounds, and a coarser stratifier gives bounds at least as wide. The width lost by a stratifier equals an expected gap between binding terms, the narrowest bounds of K strata follow from a finite search over pairs of terms, and the width lost is at most a constant times the square root of the conditional mutual information that the bottleneck minimizes. Binding terms selected on one fold and evaluated on the other give intervals without the drift of estimated maxima and minima. In a simulation study the population bounds contain the true probabilities in all 120,000 stratifier-replicate combinations. With three latent profiles the bottleneck strata attain 80\% to 85\% of the narrowing available from the full covariates at 1,500 units, also under a covariate structure that defeats outcome-blind clustering, and Euclidean clustering of the same fitted distributions gives the same widths. Intervals with selected terms cover each end of the identified set in 93.0\% to 100\% of the replicates, against 78.5\% to 100\% for plug-in intervals.}
\end{abstract}


\maketitle
\hypersetup{pdfauthor={Marek Deja},pdfsubject={}}

\section{Introduction}\label{sec:intro}

Probabilities of causation answer questions about individual units. In a civil case a patient may sue a pharmaceutical company for a side effect of one of the company's products, and the court has to judge whether the harm would have occurred without the product \citep{dawid2017probability}. When a vaccine is in short supply, a health authority wants to give the vaccine first to the people most likely to survive if vaccinated and to die if unvaccinated \citep{mueller2022causes}. \citet{pearl1999probabilities} defined three probabilities of causation on the potential outcomes of a unit, namely the probability of necessity (PN), the probability of sufficiency (PS) and the probability of necessity and sufficiency (PNS), the share of units for whom the treatment produces the outcome and the absence of the treatment prevents the outcome.

The three probabilities depend on the joint distribution of two potential outcomes of every unit, the outcome under the treatment and the outcome without the treatment, while an experiment observes only one of the two potential outcomes of a unit. Experimental data therefore identify the two margins of the joint distribution and leave the dependence between the two potential outcomes open. With both margins identified, the Fréchet--Hoeffding bounds give the sharp bounds on the joint distribution \citep{fan2010sharp}, and a partition of the units into finitely many response types turns the bounding of a causal quantity into a linear program whose optimum gives the tightest bounds compatible with the data \citep{balke1997bounds}. \citet{tian2000probabilities} derived the sharp bounds on the probabilities of causation from experimental and observational data collected in the same population, the narrowest bounds the two kinds of data allow without further assumptions. In the vaccine example of \citet{mueller2022causes}, the experimental data alone place the PNS of older patients between 20\% and 57\% and the PNS of younger patients between 10\% and 55\%, an overlap that leaves the order of vaccination undecided, and observational data on the two groups can narrow the intervals to 20\% to 40\% and 40\% to 55\%, which puts the younger patients first.

Covariates measured before the treatment narrow the bounds. For the probabilities of causation, bounds computed within every stratum of the covariates and averaged over the strata are valid and no wider than the bounds without covariates \citep{kuroki2006stratified,mueller2022causes}, and \citet{dawid2017probability} gave the analogous result for the probability of necessity. The same property holds for trimming bounds on treatment effects under sample selection \citep{lee2009training}, for bounds on principal effects \citep{long2013sharpening} and for bounds on the fraction of units harmed by a treatment, which become tighter the more finely the units are stratified by observed covariates \citep{kallus2022harm}. The guarantee is weak in the sense that averaged bounds are never wider than the bounds without covariates and are strictly narrower only under conditions on the strata \citep{long2013sharpening}, and in instrumental-variable settings a covariate affected by the instrument can make the averaged bounds wider \citep{jonzon2025adding}.

The averaging needs estimates of every experimental and observational term within every stratum, so the number of units the averaging needs grows with the number of covariate configurations. For example, in a simulation with 15 observed binary features and five million experimental and five million observational samples, only 529 of the 32,768 feature configurations appeared more than 1,300 times in both samples \citep{li2022learning}. A survey with dozens of pre-treatment questions has nearly as many configurations as respondents, and most configurations then hold a single respondent or none.

Applied work meets the dimension of the covariates by summarizing the covariates before the averaging. \citet{lee2009training} conditioned the trimming bounds on five groups formed from quintiles of a wage predicted by all baseline characteristics, \citet{long2013sharpening} proposed stratification on a dichotomized principal score, and \citet{ji2023model} described the binning of covariates as a balance between approximation quality and the number of observations per bin. The summaries come without a statement of the width lost by the summary, and \citet{levis2025covariate} left the size of the width difference between two valid sets of covariates as an open question.

A useful stratifier keeps few strata and keeps within the few strata the information that the covariates carry about the terms of the bounds. Outcome-blind clustering of the covariates, such as the $k$-means method of Lloyd in the description of \citet{arthur2007kmeans}, keeps the dominant structure of the covariates, which may be unrelated to the response to the treatment. In contrast, strata learned from the outcomes of the same units select on noise, because stratifying a randomized experiment by an outcome index fitted on the same units biases the stratum-specific effects through overfitting \citep{abadie2018endogenous}. Many poorly chosen strata also increase the variance of estimates after stratification in finite samples \citep{miratrix2013adjusting}.

The present paper learns a stratifier with few informative strata through the information bottleneck \citep{tishby1999information}, which compresses one variable into a variable with few values while keeping as much information as possible about a second variable, the relevance variable, and through the multivariate extension of the bottleneck \citep{friedman2001multivariate,slonim2006multivariate}, which compresses several groups of variables at once. The relevance variable is the outcome augmented with the design arm and the received treatment in a two-stage randomized preference design \citep{rucker1989two,long2008causal}. In the design, randomly chosen units receive an assigned treatment and the remaining units choose their treatment, so the forced-treatment arms of the design supply the experimental data and the free-choice arm supplies the observational data, both from one population. Survey experiments have used designs that randomize respondents between forced exposure and free choice, for example in an online national sample of 7,298 respondents on the effects of partisan media \citep{debenedictiskessner2019persuading}. The conditional distribution of the augmented outcome given the covariates contains every term of the bounds, and a stratifier that keeps the information about the augmented outcome therefore keeps the information the bounds need.

Compression of covariates has served causal inference before, with point identification as the target. An information bottleneck has compressed covariates into discrete clusters used as subgroups for point estimates of treatment effects under strong ignorability, with less reliable estimates as the number of clusters grew \citep{parbhoo2020information}. Representation learning has targeted the prediction of counterfactual outcomes with representations regularized toward similar distributions across treatment groups \citep{johansson2016learning}, and disentangled representations whose factors are kept independent by mutual-information estimators have targeted the dose-response function of a continuous treatment \citep{cui2026disentangled}. Bounds on probabilities of causation read the experimental and the observational distributions together, so the strata have to keep the information about the design arm, the received treatment and the outcome jointly.

Five results support the construction. Every stratifier that is a stochastic function of the pre-treatment covariates, whether learned from data or fixed in advance, gives valid bounds, and a coarser stratifier gives bounds at least as wide (Proposition~\ref{prop:nesting}). The width lost by a stratifier relative to the full covariates equals the expected gap, within the strata, between the term of a bound that binds at a covariate value and the term that binds at the stratum (Proposition~\ref{prop:loss}). The narrowest bounds that $K$ strata can give are the bounds of $K$ pairs of terms with every covariate value assigned to its narrowest pair, a finite search that reaches the bounds of the full covariates with at most 16 strata for the PNS (Proposition~\ref{prop:pairs}). The width lost is at most a constant times the square root of the conditional mutual information between the covariates and the augmented outcome given the strata, for every bound functional at once and at the best possible rate (Theorem~\ref{thm:width}), and the conditional mutual information is the quantity the bottleneck minimizes at a given level of compression. A stratifier fitted on one part of the sample and applied to the other part keeps the outcome of a unit out of the stratum of the unit, and binding terms selected on the fitting part give estimated bounds that are valid for every selection and free of the drift of estimated maxima and minima (Propositions~\ref{prop:honest} and~\ref{prop:selected}).

The two-group specification of the multivariate bottleneck compresses two groups of covariates into two latent profiles, and the combinations of the two profiles form the strata. The specification is designed for a survey of master's and doctoral students in which one group of questions describes the information reach of a student, the sources a student can draw on for a thesis, and a second group describes the information practice of the student, the use the student makes of the sources.

A simulation study with three experiments checks the claims. The population bounds of every stratifier contain the true PNS, PN and PS in all 120,000 combinations of stratifier and replicate. With three latent profiles whose dominant response type has a share of $0.85$, the bottleneck with a logistic relevance model attains 80\% to 85\% of the narrowing that the full covariates allow at 1,500 units, also when a strong covariate structure unrelated to the effect captures outcome-blind $k$-means entirely. The relevance variable carries the narrowing: $k$-means on the same fitted distributions of the augmented outcome gives widths within $0.003$ of the widths of the bottleneck, and quantile groups of a fitted effect and strata built on the exact gap of the PNS bounds give wider bounds. With an information reach profile and an information practice profile acting through separate groups of covariates, the strata of the two-group specification attain 84\% of the narrowing available from the full covariates at 1,500 units. The interval of \citet{imbens2004confidence} built on plug-in endpoints covers each end of the identified set in 78.5\% to 100\% of the replicates, pooled over the cells of a sample size, and the interval with binding terms selected out of fold in 93.0\% to 100\%, at the price of a wider interval. The bound of Theorem~\ref{thm:width} holds in every replicate and exceeds the largest possible loss in the populations of the study, so the worth of the bound lies in the rate.

\section{Probabilities of causation from combined data}\label{sec:poc}

A binary treatment $X$ with values $x$ and $x'$ and a binary outcome $Y$ with values $y$ and $y'$ define two potential outcomes for every unit, $Y_x$ under the treatment and $Y_{x'}$ without the treatment \citep{rubin1974estimating}. \citet{pearl1999probabilities} defined the three probabilities of causation on the joint distribution of the potential outcomes and the observed variables. The probability of necessity and sufficiency $\PNS = P(y_x, y'_{x'})$ is the share of units for whom the treatment produces the outcome and the absence of the treatment prevents the outcome. The probability of necessity $\PN = P(y'_{x'} \mid x, y)$ is the share of treated units with the outcome whose outcome would have been absent without the treatment. The probability of sufficiency $\PS = P(y_x \mid x', y')$ is the share of untreated units without the outcome whose outcome would have occurred under the treatment.

An experiment that assigns the treatment at random identifies the margins $P(y_x)$ and $P(y_{x'})$, and an observational study in the same population identifies the joint distribution $P(x, y)$ of the treatment the units choose and the outcome. The joint distribution of $Y_x$ and $Y_{x'}$ lies beyond both kinds of data, so the three probabilities are partially identified. \citet{tian2000probabilities} derived the sharp bounds from the two kinds of data together:
\begin{gather*}
\max\{0,\ P(y_x) - P(y_{x'}),\ P(y) - P(y_{x'}),\ P(y_x) - P(y)\} \le \PNS,\\
\PNS \le \min\{P(y_x),\ P(y'_{x'}),\ P(x, y) + P(x', y'),\\ P(y_x) - P(y_{x'}) + P(x, y') + P(x', y)\},
\end{gather*}
with
\[\max\{0, [P(y) - P(y_{x'})]/P(x, y)\} \le \PN \le \min\{1, [P(y'_{x'}) - P(x', y')]/P(x, y)\}\]
and with the $\PS$ bounds obtained by exchanging $x$ with $x'$ and $y$ with $y'$. Every term inside the maximum and the minimum is a probability, or a sum or difference of probabilities, under the experimental or the observational data, and the lower bound is the largest of the lower terms. Under monotonicity, the assumption that the treatment never prevents the outcome, the lower bounds are attained \citep{tian2000probabilities}. Bounds of the same kind follow in general from the optimization over the response types of the units, as \citet{balke1997bounds} obtained the bounds on causal effects in trials with imperfect compliance as the optimum of a linear program, and \citet{zhang2022partial} showed that the tight bounds on any counterfactual probability, the probability of necessity and sufficiency included, solve a polynomial program over causal models with finite exogenous domains. \citet{zaffalon2024efficient} mapped structural causal models with given structural equations to credal networks, computed exact counterfactual bounds for a subclass of the models and approximated the bounds of general models from inside by repeated runs of an expectation-maximization algorithm. \citet{zaffalon2023approximating} extended the algorithm to observational, randomized and selection-biased datasets fused into one model, with ranges of the PNS that typically narrowed as datasets were added. Statistical matching meets a structure of the same kind, because two variables observed in different samples that share common variables have a partially identified joint distribution, the class of distributions that extend the conditional distributions identified by the separate samples \citep{dizio2017partial}.

\citet{li2024probabilities} extended the definitions and the bounds to treatments and outcomes with several values. The extension gives bounds for quantities with one hypothetical term, presented as sharp \citep[p.~20466]{li2024probabilities}, and recursive bounds for quantities with several hypothetical terms. \citet{shu2026identification} replaced the recursive bounds with closed forms, proved the closed forms valid for every number of treatment and outcome values, verified tightness against the linear program in low dimensions and left the general proof of tightness open. For two of the single-term quantities, the probability of preservation and the probability of replacement, the lower bounds of \citet[Theorems 4 and 5]{li2024probabilities} are valid and fall below the attainable value in part of the parameter space, and the closed form of \citet[Theorem 4]{shu2026identification} with one hypothetical term gives larger lower bounds. \ref{app:lemma} proves that the lower bounds of the closed form with one hypothetical term are attained for every number of treatment and outcome values, and the software of the method implements the attained forms.

Covariates measured before the treatment narrow the bounds. \citet{kuroki2006stratified} computed the Tian--Pearl bounds within strata of observed covariates and averaged the bounds over the strata, and in a breast-cancer study of 192 patients the stratified PNS bounds were $(0.000, 0.168)$ against $(0.000, 0.237)$ without strata. \citet{mueller2022causes} showed that the Tian--Pearl bounds computed within every value $z$ of a set $Z$ of non-descendants of $X$ and averaged with weights $P(z)$ bound the population $\PNS$ and lie within the bounds without $Z$. \citet{dawid2017probability} gave the analogous result for $\PN$, and \citet{dawid2022effects} showed that with a discrete pre-exposure covariate the stratum lower bounds are attained simultaneously, so the averaged lower bound is attainable, and that the averaged lower bound equals the Tian--Pearl lower bound exactly when the risk ratios of all strata lie on the same side of 1. \citet{shu2026general} carried the averaging to the bounds for treatments and outcomes with several values. Averaged bounds are sharp if and only if the conditional bounds are uniformly sharp across the strata \citep{jonzon2025adding}. Under monotonicity the covariate-specific probability of causation is point-identified, and \citet{cuellar2020nonparametric} estimated a projection of the covariate-specific probability onto a working model with an influence-function estimator and cross-fitting. The averaging requires every term to be estimated within every stratum, from the forced-treatment arms and from the free-choice arm, and the number of strata grows exponentially with the number of covariates.

\section{The multivariate information bottleneck}\label{sec:mib}

The information bottleneck compresses a variable $C$ into a variable $T$ that keeps as much information as possible about a relevance variable $V$ at a given level of compression \citep{tishby1999information}. In the speech example of \citet{tishby1999information}, the acoustic signal is the variable to compress and the transcript of the spoken words is the relevance variable, so the compressed signal keeps the features that matter for the words and discards the rest. The encoder $P(t \mid c)$ minimizes $I(C; T) - \beta I(T; V)$, with $I$ the mutual information and $\beta > 0$ the trade-off between compression and relevance. The stationary encoders satisfy $P(t \mid c) \propto P(t) \exp\{-\beta \KL[P(V \mid c) \,\|\, P(V \mid t)]\}$, with $\KL$ the Kullback--Leibler divergence \citep{kullback1951information}, so a covariate value joins the strata whose distribution of the relevance variable resembles its own. The bottleneck generalizes rate-distortion theory, with the divergence between the distributions of the relevance variable at a value of $C$ and at a stratum as the distortion measure that emerges from the joint distribution of $C$ and $V$ \citep{tishby1999information}, and as $\beta$ grows the encoder approaches the assignment of every value of $C$ to the stratum with the smallest divergence. For finite alphabets and a known joint distribution, alternating iterations of the equations, a generalization of the Blahut--Arimoto algorithm, converge to a local solution \citep{goldfeld2020information}.

Several algorithms solve the bottleneck problem, with soft or hard assignments of the values of $C$ to the clusters. Deterministic annealing raises $\beta$ step by step, and the number of effective clusters grows through phase transitions at which clusters split, detected by keeping two perturbed copies of every cluster that separate only at a transition \citep{rose1998deterministic}. The agglomerative bottleneck starts with one cluster per value of $C$ and merges at each step the pair of clusters whose merger loses the least information about the relevance variable, a loss equal to the merged probability mass times a Jensen--Shannon divergence \citep{slonim2000agglomerative}. The sequential bottleneck keeps a fixed number of clusters, moves one value at a time into the cluster with the smallest loss of information and keeps the best of several random restarts \citep{slonim2002unsupervised}, and the deterministic bottleneck replaces $I(C; T)$ by the entropy $H(T)$, which makes the optimal encoder a hard assignment \citep{strouse2017deterministic}. The information terms of a bottleneck estimated from a sample deviate from their population values by amounts that depend on the number of clusters and not on the number of values of $C$ when the encoder is fixed independently of the sample, while the bounds grow with the number of values of $C$ when the encoder is chosen on the same sample \citep[Theorems 1 and 3]{shamir2010learning}.

\citet{friedman2001multivariate} extended the principle to several compression variables. A Bayesian network $G_{\mathrm{in}}$ states which observed variables each compression variable $T_j$ compresses, and a second Bayesian network $G_{\mathrm{out}}$ states which relations the compression variables must preserve. The Lagrangian $\mathcal{I}^{G_{\mathrm{in}}} - \beta \mathcal{I}^{G_{\mathrm{out}}}$ combines the multi-information of the two networks, each a sum of local mutual information terms between a variable and the parents of the variable in the network. The stationary encoders have the exponential form of the single bottleneck, with a distortion $d(t_j, u_j)$ that sums expected divergences over the variables that $T_j$ predicts in $G_{\mathrm{out}}$ and over the parents of $T_j$ in $G_{\mathrm{out}}$ \citep[Theorem 7.1]{friedman2001multivariate}, and asynchronous updates of the encoders converge to a stationary point \citep[Theorem 8.1]{friedman2001multivariate}. \citet{slonim2006multivariate} gave the agglomerative algorithm and the parallel, symmetric and triplet specifications of the multivariate bottleneck.

\section{Bottleneck stratification}\label{sec:method}

\subsection{Setting}

A population of units carries pre-treatment covariates $C$ with finite support. A design assigns every unit at random to one of $m + 1$ arms. The forced-treatment arm $s_j$ sets the treatment to $x_j$, for $j = 1, \dots, m$, and the free-choice arm $o$ leaves the treatment to the unit's own choice $X^{o}$. The design arm $S$ of a unit is independent of the unit's covariates, potential outcomes and choice, and the probabilities $\pi_s > 0$ of the design arms are known. The two-stage randomized preference design of \citet{rucker1989two} has the structure of the setting, and \citet{long2008causal} worked out the causal estimands of the design under self-selection. The doubly randomized preference trial of the setting randomizes every unit between the forced-treatment arms and the free-choice arm, whereas the partially randomized preference trial randomizes only the units who consent to randomization \citep{marcus2012estimating}.

The setting also assumes that the potential outcomes of a unit stay the same whether the treatment is assigned or chosen, the absence of a design effect in the terms of \citet{knox2019design}, which \citet{long2008causal} called the exclusion restriction. An effect of the mode of assignment on the outcome, for example through the satisfaction of receiving a preferred treatment, would violate the stable-unit-treatment-value assumption \citep{marcus2012estimating}. Under the two assumptions, the forced-treatment arms and the free-choice arm of a preference design with two treatments point-identify the average effect among the units who would choose each treatment \citep{knox2019design}, and the probabilities of causation remain partially identified.

The augmented outcome $V = (S, X, Y)$ records the design arm, the received treatment and the outcome of a unit. For a covariate value $c$, the conditional distribution $q_c = P(V \mid C = c)$ has the entries $q_c(s, x, y) = \pi_s P(X = x, Y = y \mid S = s, C = c)$. The forced-treatment arms give the experimental margins $P(Y_{x_j} = y \mid c) = q_c(s_j, x_j, y)/\pi_{s_j}$, and the free-choice arm gives the observational joint distribution $P(X^{o} = x, Y = y \mid c) = q_c(o, x, y)/\pi_o$. Both kinds of terms are linear functions of $q_c$, because the probabilities of the design arms are constant in $c$.

A bound functional is a pair of functions $L$ and $U$ on the set $\mathcal{D}$ of distributions of $V$ whose experimental and observational margins can come from one population, with $L(q) \le \theta \le U(q)$ for a probability $\theta$ of an event on the choice and the potential outcomes in every population with margins $q$. The Tian--Pearl bounds, the single-term bounds of \citet{li2024probabilities}, the closed forms of \citet{shu2026identification} and the exact bounds of the linear program over the joint distribution of the choice and the potential outcomes are bound functionals with $L$ convex and $U$ concave. The first three are maxima and minima of affine functions of $q$, and the optimal value of a linear program is convex in its right-hand side when the program minimizes and concave when the program maximizes. For $\PN$ and $\PS$ the functionals act on the numerators $P(y'_{x'}, x, y)$ and $P(y_x, x', y')$, and the free-choice arm identifies the denominators $P(x, y)$ and $P(x', y')$ as population quantities.

A stratifier is a stochastic map $P(t \mid c)$ from the covariates to a finite set of strata, with the stratum $T$ of a unit drawn from $P(t \mid c)$ independently of everything else given $C$. Under a stochastic map the variables $T$, $C$ and $V$ form the Markov chain $T - C - V$, and the distribution of the augmented outcome in stratum $t$ is $q_t = P(V \mid T = t) = \sum_c P(c \mid t)\, q_c$. The stratified bounds of $\theta$ are
\[
L_T = \sum_t P(t)\, L(q_t), \qquad U_T = \sum_t P(t)\, U(q_t),
\]
with $L_C$ and $U_C$ for the identity map, which keeps every covariate value as its own stratum, and $L_0$ and $U_0$ for the map onto a single stratum, which gives the bounds without covariates.

\subsection{Validity and nesting}

\begin{proposition}[Validity and nesting]\label{prop:nesting}
For every stratifier $T$ of the pre-treatment covariates, $L_T \le \theta \le U_T$. If a stratifier $T'$ is a stochastic function of $T$, then $L_{T'} \le L_T$ and $U_T \le U_{T'}$; in particular $L_0 \le L_T \le L_C \le \theta \le U_C \le U_T \le U_0$.
\end{proposition}

\begin{proof}
Within a stratum $T = t$, the design arm $S$ remains independent of the units' potential outcomes and choices, because $S$ is independent of $C$ and of the unit-level variables and $T$ depends only on $C$ and on independent noise. The forced-treatment arms therefore identify $P(Y_{x_j} \mid t)$ and the free-choice arm identifies $P(X^{o}, Y \mid t)$, so $L(q_t) \le \theta_t \le U(q_t)$ with $\theta_t$ the probability of the event in the stratum. Averaging with weights $P(t)$ gives the first statement, because $\theta = \sum_t P(t)\, \theta_t$. For the second statement, $q_{t'} = \sum_t P(t \mid t')\, q_t$ is a convex combination of points of the convex set $\mathcal{D}$, and convexity of $L$ gives $L(q_{t'}) \le \sum_t P(t \mid t')\, L(q_t)$. Multiplying by $P(t')$ and summing over $t'$ gives $L_{T'} \le L_T$, and concavity of $U$ gives the statement for the upper bounds. The chain follows with a constant $T'$ and with $T = C$, since every stratifier is a stochastic function of the covariates.
\end{proof}

The chain of Proposition~\ref{prop:nesting} contains the containment results of \citet{mueller2022causes} and \citet{shu2026general} as the comparison of the identity map with the single stratum. Every coarsening of the covariates yields valid bounds, including soft memberships and coarsenings learned from data, and refining a stratifier keeps or narrows the population bounds. The restriction to pre-treatment covariates carries the validity, in line with the widening of averaged bounds that \citet{jonzon2025adding} found for covariates affected by the instrument.

\subsection{The width lost by compression}

The Tian--Pearl bounds and the closed-form bounds for treatments and outcomes with several values are maxima and minima of finitely many affine functions of $q$. Write $L(q) = \max_i \ell_i(q)$ and $U(q) = \min_j u_j(q)$ with affine terms $\ell_i$ and $u_j$, and call a term binding at $q$ when the term attains the maximum or the minimum at $q$.

\begin{proposition}[Width lost by a stratifier]\label{prop:loss}
Let $L = \max_i \ell_i$ and $U = \min_j u_j$ with finitely many affine terms. For a stratifier $T$ of the covariates and every stratum $t$, let $\ell_{i(t)}$ and $u_{j(t)}$ be terms binding at $q_t$. Then
\begin{gather*}
L_C - L_T = \sum_t P(t)\, E\big[\max_i \ell_i(q_C) - \ell_{i(t)}(q_C) \,\big|\, T = t\big],\\
U_T - U_C = \sum_t P(t)\, E\big[u_{j(t)}(q_C) - \min_j u_j(q_C) \,\big|\, T = t\big].
\end{gather*}
The stratifier loses no width if and only if, in every stratum, the terms binding at $q_t$ are binding at $q_c$ for almost every covariate value $c$ of the stratum. The partition of the covariate values by the pair of the first binding lower term and the first binding upper term at $q_c$ attains $L_C$ and $U_C$ with at most as many strata as pairs of terms, 16 for the Tian--Pearl $\PNS$ bounds.
\end{proposition}

\begin{proof}
An affine function commutes with the average over a stratum, and $q_t = E[q_C \mid T = t]$, so $L(q_t) = \ell_{i(t)}(q_t) = E[\ell_{i(t)}(q_C) \mid T = t]$. Subtracting from $E[\max_i \ell_i(q_C) \mid T = t]$ and averaging over the strata with the weights $P(t)$ gives the identity for the lower bound, and the identity for the upper bound follows in the same way. Every conditional expectation in the two identities is the expectation of a nonnegative variable, so the width lost is zero if and only if every one of the variables is zero almost surely, which is the stated condition. In a stratum of the partition by binding terms, one lower term $\ell_i$ binds at every covariate value, so $E[L(q_C) \mid T = t] = E[\ell_i(q_C) \mid T = t] = \ell_i(q_t) \le L(q_t)$, while the convexity of $L$ gives $L(q_t) \le E[L(q_C) \mid T = t]$. The two inequalities give $L_T = L_C$, and the same argument with the binding upper term gives $U_T = U_C$.
\end{proof}

Proposition~\ref{prop:loss} states what a stratifier has to preserve. Covariate values whose bounds bind at the same terms share a stratum at no cost in width, however different their distributions of the augmented outcome, and a stratum that mixes covariate values with different binding terms loses the expected gap between the terms. The same identity with the strata in the place of the covariate values and a single stratum in the place of the stratifier shows that a stratifier narrows the bounds without covariates exactly when a term binding at the pooled distribution fails to bind in some stratum. The identity also gives the narrowest bounds that a fixed number of strata can reach.

\begin{proposition}[Narrowest bounds of $K$ strata]\label{prop:pairs}
Let $L = \max_i \ell_i$ and $U = \min_j u_j$ with finitely many affine terms, and write $w_{ij}(q) = u_j(q) - \ell_i(q)$ for the width of the pair of terms $(i, j)$ at $q$. For a set $S$ of pairs, let the stratifier $T_S$ assign every covariate value $c$ to the pair of $S$ with the smallest width at $q_c$. Every stratifier $T$ of the covariates with at most $K$ strata satisfies
\[
U_T - L_T \ge W_K, \qquad W_K = \min_{|S| \le K} E\Big[\min_{(i, j) \in S} w_{ij}(q_C)\Big],
\]
and the stratifier $T_S$ of a minimizing set $S$ attains $W_K$. In particular $W_1 = U_0 - L_0$, and $W_K = U_C - L_C$ when $K$ reaches the number of pairs of binding terms that occur among the covariate values.
\end{proposition}

\begin{proof}
Let $T$ have at most $K$ strata, and let $(i(t), j(t))$ be a pair of terms binding at $q_t$. The width of the bounds in stratum $t$ equals $w_{i(t) j(t)}(q_t)$, and the width of a pair is affine, so $U_T - L_T = \sum_{t, c} P(t, c)\, w_{i(t) j(t)}(q_c) \ge \sum_c P(c) \min_t w_{i(t) j(t)}(q_c) \ge W_K$, because the pairs of the strata form a set of at most $K$ pairs. For a set $S$ and a stratum $t$ of $T_S$ with the pair $(i, j)$, the bounds satisfy $U(q_t) - L(q_t) \le w_{ij}(q_t) = E[w_{ij}(q_C) \mid T_S = t]$, and $w_{ij}(q_c)$ is the smallest width among the pairs of $S$ at every covariate value of the stratum, so $U_{T_S} - L_{T_S} \le E[\min_{(i, j) \in S} w_{ij}(q_C)]$. For a minimizing set the two inequalities give equality. A single pair has the expected width $w_{ij}(q)$ at the pooled distribution $q$, smallest for a pair binding at $q$, which gives $W_1$. With every occurring pair of binding terms in $S$, the smallest width among the pairs of $S$ at $q_c$ is $U(q_c) - L(q_c)$, which gives $U_C - L_C$.
\end{proof}

Proposition~\ref{prop:pairs} turns the search for the narrowest $K$ strata into a choice among finitely many sets of pairs of terms, 560 sets of three pairs and 1{,}820 sets of four pairs for the Tian--Pearl $\PNS$ bounds, and the sequence $W_1 \ge W_2 \ge \dots$ shows how many strata a population needs before the bounds of the full covariates are reached. The proposition has a rate-distortion reading. A stratum is represented by a pair of terms, the distortion of representing the covariate value $c$ by the pair $(i, j)$ is the gap $w_{ij}(q_c) - [U(q_c) - L(q_c)]$, and $W_K - (U_C - L_C)$ is the smallest expected distortion of $K$ representatives. Two limits come with the exact solution. The strata of $T_S$ belong to one bound functional, so the strata that are narrowest for the $\PNS$ carry no guarantee for $\PN$, for $\PS$ or for a probability of causation of a treatment with several values. The strata also depend on the distributions $q_c$ through comparisons of terms that lie close together at many covariate values, so the assignment of a covariate value to a pair is sensitive to the error of an estimate of $q_c$, a sensitivity that the simulation study measures. A stratifier learned from data therefore needs a distortion that serves every bound functional and tolerates estimation error. The Kullback--Leibler divergence between $q_c$ and the distribution of the stratum, the distortion of the information bottleneck, is of the required kind, because the expected divergence, the conditional mutual information between the covariates and the augmented outcome given the strata, bounds the width lost under every bound functional that is Lipschitz in the sense of the next definition.

A bound functional is Lipschitz per design arm with coefficients $a_s$ and $b_s$ when
\[
|L(q) - L(q')| \le \sum_s a_s \operatorname{TV}(q_{\cdot \mid s}, q'_{\cdot \mid s}), \qquad |U(q) - U(q')| \le \sum_s b_s \operatorname{TV}(q_{\cdot \mid s}, q'_{\cdot \mid s}),
\]
with $q_{\cdot \mid s}$ the conditional distribution of $(X, Y)$ in arm $s$ of the design and $\operatorname{TV}$ the total-variation distance.

\begin{theorem}[Information bound on the width lost]\label{thm:width}
For a bound functional that is Lipschitz per design arm and for every stratifier $T$ of the covariates,
\[
0 \le (L_C - L_T) + (U_T - U_C) \le \kappa \sqrt{I(C; V \mid T)/2}, \qquad \kappa = \Big(\sum_s \frac{(a_s + b_s)^2}{\pi_s}\Big)^{1/2},
\]
with the mutual information in nats, and $I(C; V \mid T) = I(C; V) - I(T; V)$.
\end{theorem}

\begin{proof}
Since $\sum_c P(c \mid t) = 1$ and $\sum_t P(t, c) = P(c)$, the difference $L_C - L_T$ equals $\sum_{t, c} P(t, c)\, [L(q_c) - L(q_t)]$, and the same holds for $U_T - U_C$ with the sign reversed. The width lost is therefore at most $\sum_{t, c} P(t, c) \sum_s (a_s + b_s)\, \operatorname{TV}_s(c, t)$, with $\operatorname{TV}_s(c, t)$ the distance between the conditional distributions of $q_c$ and $q_t$ in arm $s$ of the design. Pinsker's inequality \citep[Theorem 7.10]{polyanskiy2025information} bounds every distance by $\sqrt{\KL_s/2}$, with $\KL_s$ the divergence between the same conditional distributions. The Cauchy--Schwarz inequality over the design arms gives $\sum_s (a_s + b_s) \sqrt{\KL_s/2} \le \kappa \sqrt{\sum_s \pi_s \KL_s / 2}$, and the chain rule of the divergence \citep[Theorem 2.15]{polyanskiy2025information}, with the same probabilities of the design arms in $q_c$ and in $q_t$, gives $\sum_s \pi_s \KL_s = \KL(q_c \,\|\, q_t)$. Jensen's inequality for the square root over the weights $P(t, c)$ moves the sum inside the root, and under the Markov chain $T - C - V$ the sum $\sum_{t, c} P(t, c)\, \KL(q_c \,\|\, q_t)$ equals $I(C; V \mid T)$. The chain rule of the mutual information \citep[Theorem 3.7]{polyanskiy2025information} gives $I(C; V \mid T) = I(C, T; V) - I(T; V) = I(C; V) - I(T; V)$, because $I(T; V \mid C) = 0$. Nonnegativity is Proposition~\ref{prop:nesting}.
\end{proof}

Every piece of the Tian--Pearl $\PNS$ bounds is a signed sum of probabilities of events, with at most one event per design arm, so the bounds are Lipschitz per design arm with $a_s = b_s = 1$ for the three arms of the binary design, the forced-$x$ arm, the forced-$x'$ arm and the free-choice arm. With equal allocation to the three arms, $\kappa = 6$. The square root cannot be replaced by a higher power of the information. Two covariate values of equal mass with uniform observational margins, $P(y_{x'}) = 1/2$ and $P(y_x) = 1/2 \pm \epsilon$, lose a width of $\epsilon$ when merged, while $I(C; V)$ is of order $\epsilon^2$.

Theorem~\ref{thm:width} is a statement of rate. The bound vanishes with the discarded information at the square-root rate, the best possible rate, and the constant makes the bound loose at moderate values of the information: with $\kappa = 6$, a bound of $0.05$ on the width lost requires $I(C; V \mid T)$ below $1.4 \times 10^{-4}$ nats. The information also counts every difference between the distributions $q_c$ within a stratum, including differences that leave the binding terms and, by Proposition~\ref{prop:loss}, the width unchanged. The conditional mutual information is a quantity of the population, computed from the distributions $q_c$ in a simulation and unknown in an application.

Theorem~\ref{thm:width} connects the width of the bounds with the bottleneck. The bottleneck with relevance variable $V$ maximizes $I(T; V)$ at a given compression $I(C; T)$, which minimizes the bound of Theorem~\ref{thm:width} at the given compression. The information serves as a surrogate for the width: a stratifier that discards little information about the augmented outcome loses little width, and a stratifier can discard much information and still keep the binding terms. The relevance variable has to be the augmented outcome, because the bound functional reads the outcome distribution of every forced-treatment arm and the joint distribution of choice and outcome in the free-choice arm, and a stratifier that preserves the distribution of one design arm only can discard the information carried by the distribution of another design arm. Proposition~\ref{prop:loss} and Theorem~\ref{thm:width} also bear on the question that \citet{levis2025covariate} left open for instrumental-variable bounds, the size of the width difference between two valid sets of covariates. In the setting of the present paper, for a stratifier that is a function of the full covariates, the difference equals the expected gap between binding terms of Proposition~\ref{prop:loss}, is no smaller than $W_K - (U_C - L_C)$ for a stratifier with $K$ strata by Proposition~\ref{prop:pairs} and is at most the constant $\kappa$ times the square root of half the information about the augmented outcome that the stratifier discards.

\subsection{Honest estimation}

\begin{proposition}[Cross-fitted stratifiers]\label{prop:honest}
Let the units be split into folds independently of the data, and let the encoder of every fold be fitted on the units outside the fold. For fixed encoders and a fixed pairing of the strata of the encoders, the stratifier that assigns a unit the memberships of the encoder of the unit's fold is a stochastic function of the covariates and of an independent fold label, so Proposition~\ref{prop:nesting} gives valid population bounds for the cross-fitted stratifier, and merging the paired strata of different fold encoders is a coarsening that keeps the bounds valid. Conditionally on the units outside a fold, the units of the fold are independent draws from the population whose strata depend on the covariates alone, and in a stratum and design arm that hold units of the fold the shares of the observed cells among the units are unbiased for the margins of the stratum under hard assignments and consistent under soft memberships.
\end{proposition}

\begin{proof}
For fixed encoders and a fixed pairing, the stratum of a unit depends on the unit's covariates and on the fold label, which is independent of the potential outcomes, the choices and the design arms. The cross-fitted stratifier therefore satisfies the conditions of Proposition~\ref{prop:nesting}, and so does every stochastic function of the cross-fitted stratifier. The folds are drawn independently of the data, so the units of a fold are independent of the units outside the fold, and the encoder of the fold is a function of the units outside the fold. Given the units outside the fold, a unit of the fold in stratum $t$ and design arm $s$ therefore has an observed cell drawn from the distribution $q_t$ of the encoder of the fold within the arm.
\end{proof}

Proposition~\ref{prop:honest} separates two properties. The population bounds of a fitted encoder are valid for every map from the covariates to strata, a map fitted on the whole sample included, and the bounds hold for every pairing of the strata of the fold encoders, so the pairing may be chosen from the data. Cross-fitting concerns the estimation of the bounds. A stratifier fitted on the same units whose margins enter the bounds would split the units along their realized outcomes, and the strata would then reflect the sampling noise of the outcomes as well as the covariates, the endogenous stratification of \citet{abadie2018endogenous}. Cross-fitting, which fits the data-dependent component on one fold, evaluates the target on the other fold and swaps the roles of the folds \citep{chernozhukov2018double}, keeps the outcome of a unit out of the stratum of the unit and gives an honest stratifier in the sense of \citet{athey2016recursive}, whose estimates within the strata behave as under a partition given in advance.

The plug-in estimates of the stratified endpoints carry a second finite-sample effect. A maximum of unbiased estimates overestimates the maximum of the means of the estimates, and a minimum underestimates the minimum, so estimated lower bounds drift upward and estimated upper bounds drift downward \citep{chernozhukov2013intersection}. Estimators of an endpoint defined by a minimum or a maximum of estimable terms are locally biased or irregular, so any bias correction leaves local bias or inflates the variance \citep{hirano2012impossibility}. The drift grows with the number of strata, because every stratum carries the estimation noise of its own terms into its own maximum and minimum, and where two terms of a bound coincide in a stratum the drift stays of the order of the standard error at every sample size. For $\PNS$ bounds conditional on covariates and fitted by a neural network, \citet{patil2026causal} corrected the drift with the precision correction of \citet{chernozhukov2013intersection}. The cross-fitted strata allow a simpler remedy, the selection of the binding terms on the units outside the fold.

\begin{proposition}[Binding terms selected out of fold]\label{prop:selected}
Let $L = \max_i \ell_i$ and $U = \min_j u_j$ with finitely many affine terms, and for every stratum $t$ of a stratifier $T$ let $\ell_{i(t)}$ be one of the lower terms and $u_{j(t)}$ one of the upper terms. Then
\[
\sum_t P(t)\, \ell_{i(t)}(q_t) \le L_T \le \theta \le U_T \le \sum_t P(t)\, u_{j(t)}(q_t),
\]
with equality on either side when the selected terms bind at their strata. If the encoder of a fold and the selected terms are functions of the units outside the fold, then, conditionally on the units outside the fold, the two outer sums are fixed numbers, and their estimates from the units of the fold, with the shares of the fold in the strata for $P(t)$ and the shares of the observed cells within the design arms for $q_t$, are smooth functions of sample proportions of independent units.
\end{proposition}

\begin{proof}
A single term of a maximum lies at or below the maximum, and a single term of a minimum at or above the minimum, so $\ell_{i(t)}(q_t) \le L(q_t)$ and $U(q_t) \le u_{j(t)}(q_t)$ in every stratum, and the average over the strata together with Proposition~\ref{prop:nesting} gives the chain. Conditionally on the units outside the fold, the encoder and the selection are fixed, the units of the fold are independent draws from the population by Proposition~\ref{prop:honest}, and each estimate is a sum over the strata of products of a stratum share and an affine function of cell shares.
\end{proof}

Bounds with binding terms selected out of fold trade the drift for a conservative shift. A term selected in error widens the bounds by the gap between the selected term and the binding term of the stratum, and the error is likely only where the gap is small relative to the noise of the selection. The estimates contain no maximum or minimum of estimated terms, so the delta method gives asymptotic normality around the two outer sums conditionally on the units outside the fold, tied terms included, and the interval of \citet{imbens2004confidence} built on the estimates of a fold covers, in large samples, every value between the two outer sums with at least the nominal probability. The set between the two outer sums contains the identified set of the full covariates. The bounds of the folds are averaged into the estimate of the method. The fold estimates depend on one another, because the selection of each fold reads the units of the other fold, so the interval of the method avoids every assumption on the dependence: with two folds, each fold interval at the level $1 - \alpha/2$ misses the true value with probability at most $\alpha/2$, and the intersection of the two fold intervals covers the true value with probability at least $1 - \alpha$ by the union bound. The construction is the stratum-level counterpart of the dual bounds of \citet{ji2023model}, whose dual variables are fitted on one fold and averaged on a separate fold and which stay valid under an inaccurate first-stage model. A small number of strata that keeps most of $I(C; V)$ limits the width lost by compression, the drift of the plug-in endpoints and the selection error of the binding terms, and the number of strata sets the balance between the width lost and the estimation error.

\section{Algorithm}\label{sec:algorithm}

\subsection{Relevance from single observations}

Every unit is observed in one design arm with one outcome, so the conditional distribution of the augmented outcome given the covariates of a unit is never observed directly. A relevance model estimates the distributions $q_c$ from the sample, and the bottleneck compresses the estimated distributions. The estimate assembles the design arms with the design probabilities, $\hat q_c(s, x, y) = \pi_s \hat P(x, y \mid s, c)$, with the distribution of the observed cell within every design arm estimated from the units of the arm. Assembling with the design probabilities keeps the random imbalance of the design arms out of the relevance variable, so $\hat q_c$ estimates the $q_c$ of Proposition~\ref{prop:loss} and Theorem~\ref{thm:width}. The paper uses two relevance models.

The logistic relevance model fits a multinomial logistic regression of the observed cell on the covariates within every design arm, the outcome in a forced-treatment arm and the pair of chosen treatment and outcome in the free-choice arm, by maximum likelihood with a ridge penalty on the coefficients. The model pools the units of a design arm over the whole covariate space, and a covariate without information about the augmented outcome receives coefficients near zero.

The nearest-neighbor relevance model smooths over neighbors in covariate space, on the premise that units close together in a suitable metric have almost the same conditional distributions \citep{cover1967nearest}. A row-stochastic smoother $W$ links every unit $i$ to the covariate configurations $a$ of the sample, and the $k$-nearest-neighbor smoother gives weight $1/k$ to each of the $k$ configurations closest to the unit. With fixed covariate weights, uniform $k$-nearest-neighbor weights estimate conditional distributions consistently when the number of neighbors grows and the share of the sample in a neighborhood vanishes \citep[Corollary 3]{stone1977consistent}, which holds for a number of neighbors equal to the square root of the number of units the encoder is fitted on, the choice of the simulation study. For a configuration $a$, the conditional distribution in arm $s$ of the design is the weighted share of each observed cell among the units of arm $s$ linked to $a$, with a small pseudo-count per cell.

Distances between units in the nearest-neighbor relevance model weight every covariate by the estimated mutual information between the covariate and the augmented outcome, reduced by the bias term $(|F| - 1)(|V| - 1)/(2n)$ for a covariate with $|F|$ values and a sample of $n$ units. The term follows from the first-order bias $-(m - 1)/(2n)$ of the plug-in entropy of a variable with $m$ cells, the basis of the Miller--Madow correction \citep[1197, 1203]{paninski2003estimation}, applied to the three entropies of $I(F; V) = H(F) + H(V) - H(F, V)$. The weighting makes the neighborhoods follow the covariates that carry information about the outcome distributions. A strong covariate structure without information about the augmented outcome, for example groups of students that differ in field or in study level and respond alike to the treatment, then gets little weight in the smoother.

\subsection{Specifications compiled from a causal graph}

A structural graph over groups of variables marks the treatment, the outcome and the design arm, and dates every group of variables before or after randomization. The graph determines the admissible inputs of every compression variable, namely the groups that are non-descendants of the treatment and of the design arm and are measured before randomization. A compression variable meant as a stratifier may take only admissible groups as inputs in $G_{\mathrm{in}}$, and $G_{\mathrm{out}}$ always contains the augmented outcome among the variables to be predicted. The single bottleneck compresses all admissible covariates into one stratifier.

The two-group specification compresses a first group of covariates, $A$, into a latent profile $T_A$ and a second group, $M$, into a latent profile $T_M$. In a survey of students who write a thesis, the setting that motivates the specification, $A$ holds the information reach features of a student, the orders in which the student ranks the sources by what each source could give and by what using each source costs, and $M$ holds the information practice features, the orders of turning to the sources, of relying on the output of the sources and of naming the sources to the supervisor. The network $G_{\mathrm{out}}$ contains $T_A \to T_M$ and $(T_A, T_M) \to V$, so the Lagrangian is
\[
I(T_A; A) + I(T_M; M) - \beta\,[\,I(T_A; T_M) + I(V; T_A, T_M)\,].
\]
The term $I(V; T_A, T_M)$ is the relevance term of Theorem~\ref{thm:width} for the strata formed by the combinations of the two profiles. The term $I(T_A; T_M)$ asks the two profiles to carry the dependence between the two groups of covariates and serves the reading of the profiles where information reach and information practice go together, while with independent groups of covariates the term is inactive. With person-unique configurations the empirical joint distribution of $A$ and $M$ is degenerate. Every value of $A$ then identifies a person, and with the person the value of $M$, so an encoder of $A$ could read $M$ through the distortion terms. The algorithm therefore defines the joint distribution through two nearest-neighbor smoothers, $P(i, a, m, v) = P(i)\, W_A[i, a]\, W_M[i, m]\, P(v \mid i)$, with $W_A$ built from the covariates of $A$, $W_M$ from the covariates of $M$ and $P(v \mid i)$ the point mass at the observed cell of unit $i$. The construction is a proper joint distribution, so the stationary conditions of \citet[Theorem 7.1]{friedman2001multivariate} and the convergence of asynchronous updates apply without change, and the profile $T_A$ receives only what the $A$-covariates of a person's neighbors carry about $V$. The two-group specification takes the design arms as observed among the neighbors of a unit, without the assembly with the design probabilities.

\subsection{Strata of pairs of terms}

Proposition~\ref{prop:pairs} gives a second way from the fitted distributions to strata, specific to one bound functional and free of a clustering step. The stratifier of pairs of terms evaluates the widths $w_{ij}$ of the 16 pairs of terms of the Tian--Pearl $\PNS$ bounds at the fitted distribution $\hat q_c$ of every unit of the fitting sample, chooses the set of $K$ pairs with the smallest mean width of the narrowest chosen pair by enumeration of all sets, and assigns a unit to the pair with the smallest width at the fitted distribution of the unit. With the distributions $q_c$ in the place of the fitted distributions, the strata are the narrowest $K$ strata of Proposition~\ref{prop:pairs}. The simulation study uses the stratifier of pairs of terms to compare the exact distortion of the $\PNS$ bounds with the distortion of the bottleneck on the same fitted distributions.

\subsection{Fitting, cross-fitting and estimation}

The encoders of the bottleneck are fitted by iterating the stationary equations with asynchronous updates and by deterministic annealing over an increasing schedule of $\beta$. At each step of the schedule every stratum is duplicated with a small perturbation, and copies that fail to separate are merged again \citep{rose1998deterministic,friedman2001multivariate}. The schedule ends at a value of $\beta$ at which the encoder assigns every covariate value to one stratum, the stratum whose distribution of the augmented outcome is nearest in the Kullback--Leibler divergence, and the solution with the chosen number of strata at the end of the schedule is kept. The number of strata is therefore the tuning constant of the algorithm, and the schedule of $\beta$ serves the optimization. A unit outside the fitting sample receives a stratum through the relevance model. With the logistic relevance model the stationary equation is evaluated at the fitted distribution of the unit, and with the nearest-neighbor relevance model the unit receives the average assignment of its nearest neighbors in the fitting sample, a soft membership.

The units are split into two folds within the design arms, and the encoder of each fold is fitted on the other fold. Two estimates of the bounds follow. The plug-in estimate matches the strata of the two fold encoders by the agreement of their memberships over all units, merges the matched strata (Proposition~\ref{prop:honest}), computes the margins of every stratum as weighted shares of the observed cells with the memberships as weights, merges strata with an expected size below five units and evaluates the bound functional at the margins of every stratum. The estimate with binding terms selected out of fold (Proposition~\ref{prop:selected}) works fold by fold. For every stratum of the encoder of a fold, the lower term and the upper term that bind at the margins of the stratum among the units outside the fold are selected, with the margins shrunk toward the pooled margins of the units outside the fold by the weight of 20 units per design arm, which steadies the selection in small strata. The selected terms are evaluated at the margins of the stratum among the units of the fold, strata with fewer than two units of a design arm in the fold are merged, and the bounds of the two folds are averaged. Standard errors come from a nonparametric bootstrap \citep{efron1979bootstrap} of units within the design arms, with memberships and selected terms held fixed. The interval of the plug-in estimate is the interval of \citet{imbens2004confidence} built on the plug-in endpoints and their standard errors. The interval of the estimate with selected terms is the intersection of the two fold intervals, each the interval of \citet{imbens2004confidence} of one fold at the level $1 - \alpha/2$.

\section{Simulation study}\label{sec:simulation}

\subsection{Design}

The simulation study answers six questions fixed in a protocol written before the study run. The first question concerns validity, namely whether the population bounds of every stratifier contain the true PNS, PN and PS. The second question concerns tightness: the width of the population bounds of each stratifier against the width without covariates, with the full covariates, with the narrowest strata of Proposition~\ref{prop:pairs} and with the true latent profiles, and the component of a stratifier that carries the narrowing. The third question concerns finite samples: how far the estimated endpoints fall from the population endpoints of the same stratifier, and how often an interval covers the true PNS and each end of the identified set. The fourth question compares the width lost to compression with the bound of Theorem~\ref{thm:width}. The fifth question asks how the strata of the two-group specification behave when an information reach profile and an information practice profile act through separate groups of covariates. The sixth question asks whether the answers hold when the latent profiles differ in the sign of the effect. The protocol, the scripts, the result tables and the software of the method are archived at Zenodo \citep{deja2026bottleneck}.

Every simulated unit belongs to a latent profile, has binary covariates whose probabilities depend on the profile, and has a response type for a binary treatment and a binary outcome. The four response types are never (the outcome is absent under both treatments), helped (the outcome occurs only under the treatment), hurt (the outcome occurs only without the treatment) and always (the outcome occurs under both). The forced-$x'$ arm, the forced-$x$ arm and the free-choice arm each have probability $1/3$. A latent profile has a binary prototype over 12 signal covariates, and a signal covariate equals the prototype with probability $0.75$, while four noise covariates have probability $1/2$ in every profile. In the free-choice arm the logit of choosing $x$ depends on the profile, rises by 1 for helped units and falls by 1 for hurt units, so the choice of treatment depends on the response type and confounds the observational data. The decoy structure adds three latent clusters, independent of the latent profiles and of the response types, and 12 further covariates that equal the prototype of the cluster with probability $0.95$, a strong covariate structure unrelated to the effect.

Experiment 1 crosses purity ($0$, $0.6$, $0.85$), decoy structure, sample size ($n = 300$, $600$, $900$, $1500$) and the number of latent profiles ($2$, $3$), with 200 replicates in each of the 48 cells and one draw of prototypes for all replicates. The purity $\rho$ sets the composition of the response types. Hurt has share $0.03$ in every profile. With $\rho > 0$ profile $r$ has a dominant type, always, helped and never in turn across the profiles, with share $\rho$, and the two other types share the rest equally. With $\rho = 0$ every profile has the composition $(0.30, 0.30, 0.03, 0.37)$ for never, helped, hurt and always. The logit of choosing $x$ runs from $-0.8$ to $0.8$ across the profiles. In the helped-dominant profile the never and the always units have equal shares and equal probabilities of choosing $x$, so the four terms of the upper PNS bound coincide in the profile, a tie that puts the estimates where estimated minima are least regular.

Experiment 2 has a binary information reach profile and a binary information practice profile, each carried by its own group of 8 signal covariates that equal the prototype with probability $0.8$ and 2 noise covariates. At purity $0.85$ the dominant response types of the four combinations of the two profiles are never, helped, helped and always. The logit of choosing $x$ runs from $-0.8$ to $0.8$ across the four combinations. The decoy structure, absent or present, joins the group of information reach covariates, and the sample size is $600$, $900$ or $1500$, with 200 replicates in each of the 6 cells. Experiment 3 has four latent profiles with the compositions $(0.08, 0.66, 0.05, 0.21)$, $(0.25, 0.05, 0.52, 0.18)$, $(0.58, 0.10, 0.06, 0.26)$ and $(0.12, 0.20, 0.10, 0.58)$ for never, helped, hurt and always and the logits $0.7$, $-0.6$, $-0.1$ and $0.3$ of choosing $x$. The effect of the treatment is negative in the second profile and positive in the other three, the binding lower terms differ across the profiles, no profile has tied terms, and the prototypes of the profiles are drawn anew in every replicate. The decoy structure is absent or present and the sample size is $600$ or $1500$, with 200 replicates in each of the 4 cells.

Twelve stratifiers enter the study. Three are references: no stratification, the true latent profiles (the oracle) and random strata. Two cluster the covariates: outcome-blind $k$-means on all covariates, run by Lloyd's algorithm from centers drawn at random among the units, the $k$-means method in the description of \citet{arthur2007kmeans}, with the best of 10 starts, and $k$-means on the covariates scaled by the relevance weights of the nearest-neighbor relevance model. Three use the bottleneck: with the nearest-neighbor relevance model, whose smoother has as many neighbors as the square root of the number of fitting units, at least 10, and a pseudo-count of $0.05$; with the same model without relevance weights; and with the logistic relevance model, whose ridge penalty is half the squared norm of the coefficients of the standardized covariates. The annealing runs over 50 values of $\beta$ from $0.5$ to $3000$ on a geometric grid. Three stratifiers start from the fitted distributions of the logistic relevance model: $k$-means with the squared Euclidean distance on the fitted distributions, the strata of pairs of terms, and quantile groups of the fitted effect $\hat P(y_x \mid c) - \hat P(y_{x'} \mid c)$, a summary of the covariates through one predicted quantity. On the same fitted distributions, the bottleneck, $k$-means and the strata of pairs of terms differ in the distortion alone: the Kullback--Leibler divergence, the squared Euclidean distance and the exact gap of the PNS bounds. Experiment 2 adds the two-group specification with two profiles from each group of covariates. The number of strata equals the number of latent profiles in Experiment 1 and is four in Experiments 2 and 3. The bottleneck without relevance weights enters the cells of Experiment 1 with three latent profiles and positive purity and the cells of Experiment 3.

Every learned stratifier is fitted with two-fold cross-fitting within the design arms. The plug-in estimate uses a pseudo-count of $0.5$ per cell in the margins of every stratum, 200 bootstrap resamples give the standard errors of both estimates, and all intervals have the nominal level 95\%.

Population quantities come from an evaluation sample of 20,000 covariate vectors drawn from the same process, with the exact conditional distributions $q_c$ of the augmented outcome. The population bounds of a stratifier built from the covariates evaluate the bound functional at $q_t$, the average of $q_c$ over the members of stratum $t$. The bounds with the full covariates average the bounds at $q_c$, the narrowest width of $K$ strata follows from Proposition~\ref{prop:pairs} by enumeration of the sets of pairs of terms, and the bounds of the oracle follow exactly from the margins of the latent profiles. The true latent profiles are no function of the covariates, so the bounds of the oracle can be narrower than the bounds with the full covariates. The identified set of the study is the interval of the bounds with the full covariates, and for the oracle the interval of the bounds of the true profiles. A population with the same distribution of all observed variables and a PNS at either end of the identified set exists, because the Tian--Pearl bounds are attained within every covariate value, and every estimator has the same distribution in a population with the PNS at an end as in the simulated population. The share of intervals that contain an end of the identified set is therefore the coverage of the interval for a PNS at the end. The conditional mutual information $I(C; V \mid T)$ is computed on the evaluation sample, and the bound of Theorem~\ref{thm:width} uses $\kappa = 6$, the constant of the Tian--Pearl PNS bounds with equal allocation to the three design arms.

\subsection{Results}

The population bounds of every stratifier contain the true PNS, the true PN and the true PS in all 120,000 combinations of stratifier and replicate of the three experiments, for the learned stratifiers, for random strata and for the oracle alike, as Proposition~\ref{prop:nesting} states. Every fit of a stratifier ended without failure.

Stratification narrows the bounds only where the binding terms differ across covariate values (Proposition~\ref{prop:loss}). At purity 0 all latent profiles share one composition of response types, and every stratifier, the full covariates and the oracle keep the width without covariates, $0.330$. With two latent profiles at purity $0.6$ and $0.85$ the lower term $P(y_x) - P(y_{x'})$ and the upper term $P(y'_{x'})$ bind in both profiles, and every stratifier, the full covariates and the oracle keep the widths without covariates, $0.215$ and $0.090$. The result belongs to the compositions of the study: the four upper terms are tied in the helped-dominant profile, so the binding upper term of the always-dominant profile binds in both profiles.

A few strata suffice in the population (Table~\ref{tab:frontier}). With three latent profiles at purity $0.85$ the width of the PNS bounds is $0.253$ without covariates and $0.130$ with the full covariates, and the narrowest two, three and four strata of Proposition~\ref{prop:pairs} give $0.144$, $0.131$ and $0.130$. In Experiment 2 the narrowest two strata give $0.103$, the width with the full covariates, and in Experiment 3 the narrowest four strata give $0.263$ against $0.262$ with the full covariates. The true latent profiles give narrower bounds than the full covariates in every setting, $0.087$ against $0.130$ at purity $0.85$, because the covariates determine the latent profile of a unit only up to an error. The covariates narrow the bounds on PN and PS as well: at purity $0.85$ the width of the PN bounds falls from $0.347$ to $0.171$ and the width of the PS bounds from $0.666$ to $0.355$.

\begin{table}[t]
\caption{Population widths of the bounds without covariates, with the narrowest strata of Proposition~\ref{prop:pairs}, with the full covariates and with the true latent profiles.}\label{tab:frontier}
\centering
\footnotesize
\setlength{\tabcolsep}{2pt}
\begin{tabular}{lrrrrrrrrrr}
\toprule
 & \multicolumn{6}{c}{PNS} & \multicolumn{2}{c}{PN} & \multicolumn{2}{c}{PS} \\
\cmidrule(lr){2-7}\cmidrule(lr){8-9}\cmidrule(lr){10-11}
Setting & 1 & 2 & 3 & 4 & Full & Profiles & 1 & Full & 1 & Full \\
\midrule
Experiment 1, purity 0.60 & 0.301 & 0.244 & 0.237 & 0.237 & 0.237 & 0.215 & 0.391 & 0.304 & 0.695 & 0.553 \\
Experiment 1, purity 0.85 & 0.253 & 0.144 & 0.131 & 0.130 & 0.130 & 0.087 & 0.347 & 0.171 & 0.666 & 0.355 \\
Experiment 2 & 0.213 & 0.103 & 0.103 & 0.103 & 0.103 & 0.088 & 0.201 & 0.091 & 0.388 & 0.199 \\
Experiment 3 & 0.345 & 0.278 & 0.268 & 0.263 & 0.262 & 0.215 & 0.436 & 0.342 & 1.000 & 0.736 \\
\bottomrule
\end{tabular}
\mstablenote{Means over 200 replicates of the cells with $n = 1500$ and without decoy structure; Experiment 1 with three latent profiles. Columns 1 to 4: narrowest bounds of one stratum, the bounds without covariates, and of two, three and four strata. Full: bounds stratified by the full covariates. Profiles: bounds stratified by the true latent profiles.}
\end{table}

With three latent profiles the learned strata narrow the bounds (Table~\ref{tab:widths}). At purity $0.85$ and $n = 1500$ the bottleneck with the logistic relevance model attains a population width of $0.149$ without decoy structure and $0.155$ with decoy structure, 85\% and 80\% of the narrowing from the bounds without covariates to the bounds with the full covariates. At $n = 300$ the widths are $0.183$ and $0.194$, 57\% and 48\% of the narrowing. At purity $0.6$ the same stratifier attains $0.255$ and $0.264$ at $n = 1500$, 71\% and 57\% of the narrowing, and $0.290$ and $0.295$ at $n = 300$, 17\% and 10\%. The strata narrow the bounds on PN and PS together with the PNS bounds: at purity $0.85$, $n = 1500$ and decoy structure the width of the PN bounds is $0.206$ and the width of the PS bounds $0.420$.

\begin{table}[t]
\caption{Population widths of the PNS bounds with three latent profiles in Experiment 1.}\label{tab:widths}
\centering
\footnotesize
\setlength{\tabcolsep}{3pt}
\begin{tabular}{lrrrrrrrr}
\toprule
 & \multicolumn{4}{c}{Purity 0.60} & \multicolumn{4}{c}{Purity 0.85} \\
\cmidrule(lr){2-5}\cmidrule(lr){6-9}
 & \multicolumn{2}{c}{No decoy} & \multicolumn{2}{c}{Decoy} & \multicolumn{2}{c}{No decoy} & \multicolumn{2}{c}{Decoy} \\
\cmidrule(lr){2-3}\cmidrule(lr){4-5}\cmidrule(lr){6-7}\cmidrule(lr){8-9}
Stratifier & 300 & 1500 & 300 & 1500 & 300 & 1500 & 300 & 1500 \\
\midrule
No strata & 0.301 & 0.301 & 0.301 & 0.301 & 0.253 & 0.253 & 0.253 & 0.253 \\
Random strata & 0.301 & 0.301 & 0.301 & 0.301 & 0.253 & 0.253 & 0.253 & 0.253 \\
$k$-means, covariates & 0.248 & 0.246 & 0.301 & 0.301 & 0.153 & 0.149 & 0.253 & 0.253 \\
$k$-means, weighted covariates & 0.283 & 0.267 & 0.289 & 0.267 & 0.195 & 0.164 & 0.200 & 0.165 \\
Bottleneck, neighbors, no weights & 0.294 & 0.277 & 0.301 & 0.290 & 0.216 & 0.192 & 0.249 & 0.215 \\
Bottleneck, neighbors & 0.290 & 0.266 & 0.293 & 0.266 & 0.198 & 0.167 & 0.199 & 0.167 \\
Bottleneck, logistic & 0.290 & 0.255 & 0.295 & 0.264 & 0.183 & 0.149 & 0.194 & 0.155 \\
$k$-means, fitted distributions & 0.291 & 0.257 & 0.296 & 0.266 & 0.184 & 0.150 & 0.197 & 0.157 \\
Pairs of terms & 0.299 & 0.285 & 0.300 & 0.295 & 0.231 & 0.179 & 0.239 & 0.195 \\
Effect quantiles & 0.301 & 0.301 & 0.301 & 0.301 & 0.253 & 0.253 & 0.253 & 0.253 \\
\midrule
Narrowest three strata & 0.237 & 0.237 & 0.237 & 0.237 & 0.131 & 0.131 & 0.131 & 0.131 \\
Full covariates & 0.237 & 0.237 & 0.237 & 0.237 & 0.130 & 0.130 & 0.130 & 0.130 \\
True latent profiles & 0.215 & 0.215 & 0.215 & 0.215 & 0.087 & 0.087 & 0.087 & 0.087 \\
\bottomrule
\end{tabular}
\mstablenote{Means over 200 replicates per cell; columns give the sample size $n$. Purity: share of the dominant response type in every latent profile. Decoy: three latent clusters of 12 further covariates, independent of the latent profiles and of the response types. $k$-means, covariates: outcome-blind clusters. Weighted covariates: covariates scaled by the relevance weights of the nearest-neighbor relevance model. Neighbors and logistic: relevance model of the bottleneck. Fitted distributions: fitted distributions of the logistic relevance model, which the strata of pairs of terms and the quantile groups of the fitted effect use as well. Narrowest three strata: Proposition~\ref{prop:pairs}.}
\end{table}

Outcome-blind $k$-means depends on the geometry of the covariates. Without decoy structure the latent profiles are the only cluster structure of the covariates, and at purity $0.85$ $k$-means attains $0.153$ at $n = 300$ and $0.149$ at $n = 1500$, the narrowest bounds of all learned stratifiers at $n = 300$. With decoy structure the clusters of $k$-means follow the decoy clusters, which are independent of the response types, and the width stays at $0.253$, the width without covariates. Random strata keep the width without covariates in every cell.

The relevance weights carry the gain of the bottleneck with the nearest-neighbor relevance model. At purity $0.85$ and $n = 1500$ the bottleneck with the nearest-neighbor model attains $0.167$ with and without decoy structure, and the same bottleneck without relevance weights attains $0.192$ and $0.215$. $k$-means on the covariates scaled by the relevance weights attains $0.164$ and $0.165$, and the mean paired difference between the widths of the bottleneck and of the weighted $k$-means lies between $-0.002$ and $0.007$ in the 16 cells with three latent profiles and positive purity, with standard errors of at most $0.0013$. On the weighted covariates the compression of the bottleneck and the clustering of $k$-means give the same widths.

On the fitted distributions of the logistic relevance model the three distortions keep one order in all 16 cells. The bottleneck is narrower than $k$-means on the fitted distributions by $0.0002$ to $0.0028$ and narrower than the strata of pairs of terms by $0.005$ to $0.052$. The exact distortion of the PNS bounds, which gives the narrowest strata at the distributions $q_c$, gives the widest of the three sets of strata at the fitted distributions. The logistic relevance model gives narrower bounds than the nearest-neighbor model at purity $0.85$, by $0.004$ to $0.018$, and at purity $0.6$ the difference between the two models lies between $-0.004$ and $0.011$. Quantile groups of the fitted effect keep the width without covariates, $0.301$ at purity $0.6$ and $0.253$ at purity $0.85$: the profiles with the dominant types always and never have equal effects of the treatment and different binding upper terms, and strata formed from the effect leave the two profiles together.

With an information reach profile and an information practice profile on separate groups of covariates, the two-group specification attains $0.138$, $0.125$ and $0.120$ at $n = 600$, $900$ and $1500$ without decoy structure and $0.140$, $0.127$ and $0.121$ with decoy structure, against $0.212$ to $0.213$ without covariates and $0.103$ with the full covariates (Table~\ref{tab:twogroup}). With decoy structure the two-group specification gives the narrowest bounds of all learned stratifiers at every sample size, narrower than the bottleneck with the logistic relevance model by $0.010$, $0.011$ and $0.006$. Without decoy structure the two-group specification and the bottleneck with the logistic model differ by less than $0.005$ in either direction, and outcome-blind $k$-means, with $0.120$ and $0.118$, gives the narrowest bounds at $n = 600$ and $900$ and stays at $0.199$ to $0.200$ with decoy structure. The learned profiles agree with the true profiles more closely as the sample grows. Without decoy structure the adjusted mutual information \citep{vinh2010information}, with the arithmetic mean of the two entropies as normalization, rises from $0.38$ at $n = 600$ to $0.49$ at $n = 1500$ for the information reach profile and from $0.39$ to $0.52$ for the information practice profile.

\begin{table}[t]
\caption{Population widths of the PNS bounds with an information reach profile and an information practice profile on separate groups of covariates in Experiment 2.}\label{tab:twogroup}
\centering
\footnotesize
\setlength{\tabcolsep}{3pt}
\begin{tabular}{lrrrrrr}
\toprule
 & \multicolumn{3}{c}{No decoy} & \multicolumn{3}{c}{Decoy} \\
\cmidrule(lr){2-4}\cmidrule(lr){5-7}
Stratifier & 600 & 900 & 1500 & 600 & 900 & 1500 \\
\midrule
No strata & 0.212 & 0.212 & 0.213 & 0.212 & 0.212 & 0.212 \\
$k$-means, covariates & 0.120 & 0.118 & 0.117 & 0.200 & 0.199 & 0.199 \\
$k$-means, weighted covariates & 0.142 & 0.135 & 0.129 & 0.143 & 0.135 & 0.130 \\
Bottleneck, neighbors & 0.143 & 0.139 & 0.134 & 0.146 & 0.140 & 0.135 \\
Bottleneck, logistic & 0.138 & 0.130 & 0.117 & 0.150 & 0.139 & 0.127 \\
$k$-means, fitted distributions & 0.139 & 0.128 & 0.116 & 0.150 & 0.140 & 0.125 \\
Pairs of terms & 0.166 & 0.146 & 0.127 & 0.181 & 0.168 & 0.142 \\
Effect quantiles & 0.212 & 0.212 & 0.213 & 0.212 & 0.212 & 0.212 \\
Two-group specification & 0.138 & 0.125 & 0.120 & 0.140 & 0.127 & 0.121 \\
\midrule
Narrowest four strata & 0.103 & 0.103 & 0.103 & 0.103 & 0.103 & 0.103 \\
Full covariates & 0.103 & 0.103 & 0.103 & 0.103 & 0.103 & 0.103 \\
True latent profiles & 0.088 & 0.088 & 0.088 & 0.088 & 0.088 & 0.088 \\
\bottomrule
\end{tabular}
\mstablenote{Means over 200 replicates per cell; columns give the sample size $n$; four strata. Decoy: three latent clusters of 12 further covariates joined to the group of information reach covariates. Two-group specification: two profiles from each group of covariates. The other rows as in Table~\ref{tab:widths}.}
\end{table}

With latent profiles of both signs of the effect the full covariates narrow the bounds from $0.345$ to $0.262$, and the learned strata recover a part of the narrowing (Table~\ref{tab:mixed}). At $n = 1500$ without decoy structure the bottleneck with the logistic relevance model attains $0.298$ and $k$-means on the fitted distributions $0.297$, 57\% and 58\% of the narrowing, the weighted $k$-means $0.309$ and the bottleneck with the nearest-neighbor model $0.312$. With decoy structure the four stratifiers lie between $0.308$ and $0.313$ at $n = 1500$, and at $n = 600$ the weighted $k$-means gives the narrowest bounds of the four, $0.318$. Outcome-blind $k$-means attains $0.276$ at $n = 1500$ without decoy structure and $0.336$ with decoy structure. $k$-means on the fitted distributions is narrower than the bottleneck with the logistic model by $0.001$ to $0.002$ in the four cells, the reverse of the order in Experiment 1. The narrowing comes from the upper bound. At $n = 1500$ without decoy structure the bottleneck with the logistic model moves the upper bound from $0.510$ to $0.468$, against $0.442$ with the full covariates, and the lower bound from $0.165$ to $0.170$, against $0.180$.

\begin{table}[t]
\caption{Population PNS bounds with four latent profiles of both signs of the effect in Experiment 3.}\label{tab:mixed}
\centering
\footnotesize
\setlength{\tabcolsep}{3pt}
\begin{tabular}{lrrrrrr}
\toprule
 & \multicolumn{4}{c}{Width} & \multicolumn{2}{c}{Endpoints} \\
\cmidrule(lr){2-5}\cmidrule(lr){6-7}
 & \multicolumn{2}{c}{No decoy} & \multicolumn{2}{c}{Decoy} & & \\
\cmidrule(lr){2-3}\cmidrule(lr){4-5}
Stratifier & 600 & 1500 & 600 & 1500 & Lower & Upper \\
\midrule
No strata & 0.345 & 0.345 & 0.345 & 0.345 & 0.165 & 0.510 \\
Random strata & 0.345 & 0.345 & 0.345 & 0.345 & 0.165 & 0.510 \\
$k$-means, covariates & 0.281 & 0.276 & 0.337 & 0.336 & 0.172 & 0.448 \\
$k$-means, weighted covariates & 0.313 & 0.309 & 0.318 & 0.311 & 0.166 & 0.476 \\
Bottleneck, neighbors, no weights & 0.326 & 0.318 & 0.337 & 0.330 & 0.165 & 0.483 \\
Bottleneck, neighbors & 0.321 & 0.312 & 0.324 & 0.313 & 0.166 & 0.478 \\
Bottleneck, logistic & 0.317 & 0.298 & 0.328 & 0.311 & 0.170 & 0.468 \\
$k$-means, fitted distributions & 0.315 & 0.297 & 0.327 & 0.308 & 0.170 & 0.467 \\
Pairs of terms & 0.332 & 0.323 & 0.338 & 0.332 & 0.165 & 0.487 \\
Effect quantiles & 0.334 & 0.328 & 0.340 & 0.334 & 0.169 & 0.498 \\
\midrule
Narrowest four strata & 0.263 & 0.263 & 0.263 & 0.263 & & \\
Full covariates & 0.262 & 0.262 & 0.262 & 0.262 & 0.180 & 0.442 \\
True latent profiles & 0.215 & 0.215 & 0.215 & 0.215 & 0.198 & 0.412 \\
\bottomrule
\end{tabular}
\mstablenote{Means over 200 replicates per cell, with the prototypes of the latent profiles drawn anew in every replicate; width columns give the sample size $n$; four strata. Endpoints: lower and upper population bound at $n = 1500$ without decoy structure. The rows as in Table~\ref{tab:widths}.}
\end{table}

The width lost by a stratifier relative to the full covariates stays below the bound of Theorem~\ref{thm:width} in every replicate of the three experiments, and the largest ratio of the loss to the bound is $0.083$. The numerical value of the bound is uninformative in the settings of the study: the smallest bound over all replicates is $0.170$, and the largest possible loss, the width without covariates minus the width with the full covariates, is at most $0.125$. The bound of Theorem~\ref{thm:width} holds for every bound functional at once and gives the rate at which the loss vanishes with the information lost, and a numerical statement about the loss in a given population needs the exact identity of Proposition~\ref{prop:loss}.

The two estimates err in opposite directions in small samples (Table~\ref{tab:finite}). The plug-in estimate is narrower than the population bounds of the same stratifier, because an estimated maximum of terms lies above and an estimated minimum of terms lies below the population value on average. With three latent profiles, purity $0.85$ and decoy structure, the bottleneck with the logistic model has a population width of $0.194$ and a mean plug-in width of $0.157$ at $n = 300$, and $0.155$ and $0.147$ at $n = 1500$. The estimate with selected terms is wider than the population bounds, $0.260$ at $n = 300$ and $0.174$ at $n = 1500$: a term selected on one half of the sample differs at times from the binding term, and every term other than the binding term gives a wider, valid bound.

\begin{table}[t]
\caption{Mean widths of the population bounds, of the two estimates and of the two 95\% intervals in Experiment 1 with three latent profiles, purity $0.85$ and decoy structure.}\label{tab:finite}
\centering
\footnotesize
\setlength{\tabcolsep}{3pt}
\begin{tabular}{lrrrr}
\toprule
 & 300 & 600 & 900 & 1500 \\
\midrule
\multicolumn{5}{l}{No strata} \\
\quad Population bounds & 0.253 & 0.253 & 0.253 & 0.253 \\
\quad Plug-in estimate & 0.250 & 0.251 & 0.248 & 0.251 \\
\quad Estimate with selected terms & 0.313 & 0.278 & 0.262 & 0.258 \\
\quad Plug-in interval & 0.427 & 0.380 & 0.356 & 0.335 \\
\quad Interval with selected terms & 0.543 & 0.441 & 0.399 & 0.362 \\
\multicolumn{5}{l}{Bottleneck, neighbors} \\
\quad Population bounds & 0.199 & 0.181 & 0.173 & 0.167 \\
\quad Plug-in estimate & 0.180 & 0.170 & 0.168 & 0.162 \\
\quad Estimate with selected terms & 0.262 & 0.213 & 0.196 & 0.180 \\
\quad Plug-in interval & 0.332 & 0.277 & 0.256 & 0.231 \\
\quad Interval with selected terms & 0.468 & 0.351 & 0.309 & 0.264 \\
\multicolumn{5}{l}{Bottleneck, logistic} \\
\quad Population bounds & 0.194 & 0.177 & 0.166 & 0.155 \\
\quad Plug-in estimate & 0.157 & 0.161 & 0.154 & 0.147 \\
\quad Estimate with selected terms & 0.260 & 0.228 & 0.199 & 0.174 \\
\quad Plug-in interval & 0.318 & 0.278 & 0.249 & 0.221 \\
\quad Interval with selected terms & 0.477 & 0.382 & 0.322 & 0.265 \\
\bottomrule
\end{tabular}
\mstablenote{Means over 200 replicates per cell; columns give the sample size $n$. Plug-in estimate: bound functional at the pooled margins of the aligned fold strata. Estimate with selected terms: binding terms selected out of fold. Plug-in interval: interval of \citet{imbens2004confidence} on the plug-in endpoints. Interval with selected terms: intersection of the fold intervals at the level $0.975$.}
\end{table}

The interval of the plug-in estimate covers the true PNS in at least 93.0\% of the replicates of every cell and stratifier of Experiment 1 and loses the coverage at the ends of the identified set (Table~\ref{tab:coverage}). In the cells where the full covariates leave the bounds unchanged, where terms of the upper bound are close or tied, the plug-in interval of the bottleneck with the logistic model covers the upper end in 83.6\% of the replicates at $n = 300$ and in 91.0\% at $n = 1500$, and the lowest share of a single cell and stratifier is 66.5\%. The plug-in upper endpoint, a minimum of estimated terms, lies below the population endpoint on average, and an interval built on a biased endpoint loses the coverage of a PNS at the end. The interval of the estimate with selected terms covers each end in 96.0\% to 100\% of the replicates for the bottleneck with the logistic model and in 93.0\% to 100\% for every other stratifier, pooled over the cells of a sample size. The lowest shares of a single cell and stratifier are 90.0\% at the lower end and 92.0\% at the upper end, where the Monte Carlo standard error of a cell of 200 replicates is $1.5$ percentage points at a coverage of 95\%.

\begin{table}[t]
\caption{Coverage of the 95\% intervals in Experiment 1: shares of replicates whose interval contains the true PNS, the lower end and the upper end of the identified set.}\label{tab:coverage}
\centering
\footnotesize
\setlength{\tabcolsep}{3pt}
\begin{tabular}{lrrrrrrrr}
\toprule
 & \multicolumn{4}{c}{Bounds unchanged by covariates} & \multicolumn{4}{c}{Bounds narrowed by covariates} \\
\cmidrule(lr){2-5}\cmidrule(lr){6-9}
 & 300 & 600 & 900 & 1500 & 300 & 600 & 900 & 1500 \\
\midrule
\multicolumn{9}{l}{Plug-in interval, no strata} \\
\quad True PNS & 0.978 & 0.987 & 0.995 & 0.998 & 0.969 & 0.984 & 0.995 & 0.996 \\
\quad Lower end & 0.938 & 0.941 & 0.949 & 0.943 & 0.931 & 0.943 & 0.940 & 0.951 \\
\quad Upper end & 0.926 & 0.939 & 0.949 & 0.940 & 0.999 & 1.000 & 1.000 & 1.000 \\
\multicolumn{9}{l}{Plug-in interval, bottleneck} \\
\quad True PNS & 0.971 & 0.988 & 0.995 & 0.998 & 0.963 & 0.984 & 0.994 & 0.995 \\
\quad Lower end & 0.929 & 0.943 & 0.944 & 0.938 & 0.906 & 0.920 & 0.929 & 0.938 \\
\quad Upper end & 0.836 & 0.881 & 0.909 & 0.910 & 0.976 & 0.988 & 0.991 & 0.991 \\
\multicolumn{9}{l}{Interval with selected terms, no strata} \\
\quad True PNS & 0.971 & 0.981 & 0.987 & 0.993 & 0.966 & 0.971 & 0.981 & 0.994 \\
\quad Lower end & 0.945 & 0.940 & 0.947 & 0.946 & 0.945 & 0.936 & 0.930 & 0.956 \\
\quad Upper end & 0.960 & 0.961 & 0.958 & 0.954 & 0.996 & 0.999 & 1.000 & 1.000 \\
\multicolumn{9}{l}{Interval with selected terms, bottleneck} \\
\quad True PNS & 0.979 & 0.990 & 0.992 & 0.997 & 0.976 & 0.986 & 0.986 & 0.996 \\
\quad Lower end & 0.961 & 0.965 & 0.968 & 0.969 & 0.965 & 0.960 & 0.966 & 0.960 \\
\quad Upper end & 0.972 & 0.981 & 0.973 & 0.978 & 1.000 & 0.998 & 0.998 & 0.994 \\
\bottomrule
\end{tabular}
\mstablenote{Columns give the sample size $n$. Bounds unchanged by covariates: the 32 cells with purity 0 or with two latent profiles, 1,600 replicates per entry. Bounds narrowed by covariates: the 16 cells with three latent profiles and positive purity, 800 replicates per entry. Bottleneck: the bottleneck with the logistic relevance model. Identified set: interval of the population bounds with the full covariates.}
\end{table}

The coverage has a price in width (Table~\ref{tab:finite}). With three latent profiles, purity $0.85$ and decoy structure, the interval with selected terms of the bottleneck with the logistic model has a mean width of $0.477$ at $n = 300$ and $0.265$ at $n = 1500$, against $0.318$ and $0.221$ for the plug-in interval of the same strata. The plug-in interval without strata has a mean width of $0.427$, $0.380$, $0.356$ and $0.335$ at $n = 300$, $600$, $900$ and $1500$. The interval with selected terms of the bottleneck with the nearest-neighbor model is wider at $n = 300$, with $0.468$, and narrower from $n = 600$ on, with $0.351$, $0.309$ and $0.264$, and the interval with selected terms of the bottleneck with the logistic model is narrower from $n = 900$ on. In the condition of the study with the largest narrowing, learned strata shorten an interval with coverage at the ends from 600 to 900 units on, and the population bounds narrow at every sample size.

In Experiment 2 the interval with selected terms of the two-group specification covers the lower end of the identified set in 94.0\% to 95.8\% and the upper end in 98.5\% to 100\% of the replicates, pooled over the two decoy conditions. In Experiment 3 the interval with selected terms covers each end in at least 97.0\% of the replicates for every stratifier, pooled over the two decoy conditions. An interval built on the averages of the fold endpoints, with standard errors that treat the two folds as independent, is 6\% narrower on average than the intersection of the fold intervals, and the lowest share of replicates of a single cell and stratifier in which the interval covers an end of the identified set is 90.0\%.

\section{Discussion}\label{sec:discussion}

\subsection{Reading of the results}

The containment results of \citet{mueller2022causes}, \citet{dawid2017probability} and \citet{shu2026general} compare the bounds with all covariates to the bounds without covariates. Proposition~\ref{prop:nesting} places every stratifier between the two, and the simulation confirms the ordering in every replicate, for the learned stratifiers, for random strata and for the true latent profiles alike. As a consequence, validity asks of the stratifier only a dependence on pre-treatment covariates and on an independent fold label, and the choice among stratifiers becomes a question of tightness and of estimation error.

Tightness depends on the response types more than on the covariates. Stratification narrows the population bounds only where the strata differ in which term of a bound attains the extreme (Proposition~\ref{prop:loss}), and the simulation shows both cases. Two latent profiles whose bounds share their binding terms keep the width without covariates, with the true profiles and with every learned stratifier, while three latent profiles with different binding terms narrow the bounds from $0.253$ to $0.130$ with the full covariates. The result for two latent profiles belongs to the compositions of the study, whose helped-dominant profile has tied upper terms, and two latent profiles with different binding terms narrow the bounds as three do. Proposition~\ref{prop:pairs} adds the number of strata to the picture. Three strata chosen with knowledge of the population reach $0.131$, so the task of a stratifier in a sample is the estimation of a partition with few cells, and the width of learned strata in excess of the narrowest strata is the error of the estimated partition.

The relevance variable carries the narrowing, and the distortion measure matters little. Three comparisons of the study support the reading. Clusters of the covariates narrow the bounds when the geometry of the covariates matches the latent profiles and keep the width without covariates when a stronger structure unrelated to the effect dominates the geometry, while every stratifier that reads the augmented outcome keeps a gain under the unrelated structure. With the nearest-neighbor relevance model the relevance weights carry the gain, because $k$-means on the weighted covariates matches the bottleneck and the bottleneck without the weights loses a large part of the gain. On the fitted distributions of the logistic relevance model the Kullback--Leibler divergence of the bottleneck and the squared Euclidean distance of $k$-means give widths within $0.003$ of one another in every cell of the three experiments, with the bottleneck narrower in Experiment 1 and $k$-means narrower in Experiment 3. The method therefore consists of the augmented outcome as the relevance variable and of a relevance model that predicts the augmented outcome from the covariates, and the information bottleneck is the clustering rule for which Theorem~\ref{thm:width} bounds the loss of every bound functional at once.

The exact distortion of one bound functional, which defines the narrowest strata in the population, gives wider bounds than both clustering rules when the distributions are fitted. A possible reason lies in the form of the exact distortion. The assignment of a unit to a pair of terms changes where two terms of a bound cross, so a small error of a fitted distribution near a crossing moves the unit to another pair, while the Kullback--Leibler divergence and the Euclidean distance change smoothly with the fitted distribution. The study measures the result and leaves the reason untested. Summaries of the covariates through one predicted quantity, as in the quintiles of a predicted outcome \citep{lee2009training} or in a dichotomized principal score \citep{long2013sharpening}, can miss the narrowing altogether: quantile groups of the fitted effect keep the width without covariates in the populations of the study in which two latent profiles have equal effects and different binding terms.

With an information reach profile and an information practice profile carried by separate groups of covariates, the strata of the two-group specification are as narrow as the strata of a single bottleneck without decoy structure and the narrowest of all learned strata with decoy structure in one of the two groups, and the learned profiles agree more closely with the true profiles as the sample grows. The two profiles add a reading to the strata, because each stratum is a combination of a level of information reach and a level of information practice.

The plug-in estimate and the plug-in interval are optimistic. The estimated bounds are narrower than the population bounds, as the drift of plug-in maxima and minima predicts \citep{chernozhukov2013intersection}. The interval built on the plug-in endpoints covers the true PNS of the simulated populations, which lies in the interior of the identified set, and covers a PNS at an end of the identified set in as few as 78.5\% of the replicates pooled over the cells of a sample size. A partially identified parameter can lie anywhere in the identified set, so an interval for the parameter has to cover the ends as well \citep{imbens2004confidence}. The estimate with binding terms selected out of fold replaces the drift by a conservative shift, and the intersection of the fold intervals covers each end in 93.0\% to 100\% of the replicates pooled over the cells of a sample size, with single cells of 200 replicates down to 90.0\%. The lowest cells occur without strata as well as with strata. The coverage has a price in width, and the price sets the sample size at which learned strata pay in the length of the interval: in the condition of the study with the largest narrowing, intervals with learned strata are shorter than the plug-in interval without strata from 600 to 900 units on and longer at 300 units. Below a few hundred units per design arm the bounds without covariates are the better report, and the learned strata describe the heterogeneity of the population bounds.

The bound of Theorem~\ref{thm:width} holds in every replicate and exceeds the largest possible loss in every replicate, so the bound gives no numerical information in the study. The theorem bounds the worst case over all populations with the same information, and the two-point example of the theory attains the square-root order of the bound. The exact identity of Proposition~\ref{prop:loss} and the finite search of Proposition~\ref{prop:pairs} are the tools for a given population, and the two tools need the conditional distributions $q_c$, which a sample provides only through a fitted model.

Estimators of covariate-assisted bounds without strata fit functions of the covariates by regression. For bounds on average treatment effects from instrumental variables, \citet{levis2025covariate} proposed estimators based on influence functions that attain parametric rates under a margin condition. For the fraction of units harmed by a treatment, the inference of \citet{kallus2022harm} stays consistent when some of the learned functions are wrong. The dual bounds of \citet{ji2023model} wrap around any estimates of the conditional distributions of the potential outcomes and stay valid in randomized experiments when the estimates are inaccurate, and the authors note the efficiency that a binning of covariates can lose. Each of the three estimators targets one estimand with its own fitted functions. Strata give one partition of the units for every bound functional of the design, the bounds on PNS, PN and PS in the study, are valid without a rate or a margin condition on the fitted model (Proposition~\ref{prop:nesting}), reduce inference to proportions within strata (Proposition~\ref{prop:selected}) and describe profiles of units that a report of a survey can name. The price is the width between the bounds of the strata and the bounds of the full covariates, which Propositions~\ref{prop:loss} and~\ref{prop:pairs} measure. The structural approach of \citet{zaffalon2023approximating} fuses experimental and observational datasets in one structural causal model and approximates the range of the PNS from inside, where the bounds of the present method are closed-form functionals evaluated within strata. The study compares stratifiers with one another, and a comparison of strata with the estimators without strata and with the structural approach on common data is open.

Incompatible margins are a second effect of finite samples. \citet[Example 7]{zaffalon2024efficient} showed that data generated by a structural causal model with restricted mechanisms can be incompatible with the model, so that bounds derived from the factorization of the observed distribution exclude the true PNS of the model, and asked for probabilistic guarantees on bounds computed from finite samples. The bounds of the present method admit every response type, and the compatibility that matters lies between the two kinds of margins: in the population, consistency, $Y = Y_x$ for the units that choose $x$, gives $P(y_x) \ge P(x, y)$ for every value of the treatment and of the outcome, and margins estimated within a small stratum can violate the inequality. In the simulation, incompatible strata are most frequent where the strata are small and the response types pure. At purity $0.85$ with three latent profiles and without decoy structure, a replicate holds on average $0.490$ incompatible strata of the bottleneck with the nearest-neighbor model and $1.215$ incompatible strata of the true latent profiles at $n = 300$, and $0.025$ and $0.400$ at $n = 1500$. The bound functional is evaluated at the estimated margins of every stratum, and in the cell with $n = 300$ the interval with selected terms covers the lower and the upper end of the identified set in $94.5\%$ and $100\%$ of the replicates with the bottleneck strata and in $95.0\%$ and $96.0\%$ with the true latent profiles. The interval keeps the coverage where the estimated margins of single strata are incompatible.

\subsection{Limitations}

The evidence is simulated, and an application of the method to field data is open. The simulation covers binary treatments and outcomes. The bound functionals for treatments and outcomes with several values are verified against the exact linear program in the tests of the software and enter the theory through Proposition~\ref{prop:nesting} and Theorem~\ref{thm:width}, with their own Lipschitz constants. The data-generating processes make the covariates informative only through latent profiles, with covariates that are independent binary variables given the profile, a structure that suits the logistic relevance model, and covariates with graded heterogeneity or with interactions would test both relevance models more severely. The study compares stratifiers and leaves out the estimators of covariate-assisted bounds without strata. The number of strata equals the number of latent profiles in the study, and a rule that chooses the number of strata from a sample is open. The two groups of covariates of Experiment 2 are independent, so the term $I(T_A; T_M)$ of the two-group specification is inactive in the study and the effect of the term on the strata is untested.

The interval with selected terms rests on the normal approximation of proportions within folds, strata and design arms and on a bootstrap that holds the fitted encoders and the selected terms fixed, which Proposition~\ref{prop:selected} justifies conditionally on the units outside the fold. The coverage of single cells falls to 90.0\%, and the approximation is least accurate in strata with response types of high purity, whose cell proportions lie near 0 and 1. Two folds leave half of the sample for the selection and half for the evaluation, and the union bound adds width where the two folds agree, so the interval is wider than an interval for strata given in advance. Estimators of an endpoint defined by a maximum or a minimum face the trade-off between bias and variance that \citet{hirano2012impossibility} proved unavoidable, and the selection out of fold takes the conservative side of the trade-off. The interval of \citet{imbens2004confidence} assumes superefficient estimation of the length of the identified interval and can lose uniform coverage near point identification, which the interval of \citet{stoye2009more} avoids. The validity of the bounds rests on the absence of a design effect \citep{knox2019design}, an assumption that an effect of the mode of assignment on the outcome breaks \citep{marcus2012estimating}, and a violation of the assumption affects the bounds of every stratifier alike. Equal allocation to the three design arms gives the constant $\kappa = 6$, and the constant grows as the probability of the smallest design arm falls, so the bound of Theorem~\ref{thm:width} loosens in unbalanced designs. The gains of stratification require response types that differ across latent profiles in their binding terms, a property of the population that a stratifier can reveal and leave unchanged. The choice of the covariates entering the bottleneck stays with the analyst, and a formal rule for selecting covariates with valid inference after the selection, which \citet{long2013sharpening} named as a task for future research, lies outside the method.

\section{Conclusions}\label{sec:conclusions}

Pre-treatment covariates narrow the bounds on probabilities of causation from combined experimental and observational data, and the narrowing fits into a few strata. The narrowest bounds of $K$ strata are the bounds of $K$ pairs of binding terms, and three or four strata reach the bounds of the full covariates in the simulated populations. Validity holds for every stratifier of the pre-treatment covariates, the width lost by compression equals an expected gap between binding terms and is bounded by the information about the augmented outcome that the strata discard, and binding terms selected out of fold give intervals that keep the coverage at the ends of the identified set.

In the simulation the bottleneck with a logistic relevance model attains 80\% to 85\% of the narrowing available from the full covariates with three latent profiles of high purity at 1,500 units and keeps the gain when a strong covariate structure unrelated to the effect defeats outcome-blind clustering, and the two-group specification attains 84\%. Clustering of the same fitted distributions with the Euclidean distance gives the same widths, so the augmented outcome as the relevance variable is the component that carries the gain, and the information bottleneck is the clustering rule with a bound on the loss for every bound functional. Intervals with coverage at the ends of the identified set are wider than plug-in intervals, and learned strata shorten the intervals from 600 to 900 units on. The method suits survey experiments with a randomized preference design, in which many pre-treatment questions describe each respondent and a few strata have to carry what the questions say about the response to the treatment.

\appendix

\section{Attained lower bounds for two single-term probabilities of causation}\label{app:lemma}

Let the treatment $X$ take the values $x_1, \dots, x_m$ and the outcome $Y$ the values $y_1, \dots, y_n$, and write $P(y_{i, x_j}) = P(Y_{x_j} = y_i)$ for the experimental margins. \citet[Theorems 4 and 5]{li2024probabilities} bound the probability of preservation $P(Y_{x_j} = y_i, Y = y_i)$ from below by $\max\{P(x_j, y_i),\ P(y_{i, x_j}) + P(y_i) - 1\}$, and the probability of replacement $P(Y_{x_j} = y_i, Y = y_k)$, $k \ne i$, by the largest of $0$, $P(y_{i, x_j}) + P(y_k) - 1$ and a sum over the treatment values $x_p \ne x_j$ of per-value terms. Both lower bounds are valid. The closed form of \citet[Theorem 4]{shu2026identification} for the probability of replacement with several hypothetical terms, taken with one hypothetical term, gives the two lower bounds of Lemma~\ref{lem:sharp}. The closed form is stated for equal numbers of treatment and outcome values with aligned indices, extends to other cases by a relabeling of the values \citep[Theorems 6 and 7]{shu2026identification} and comes with a proof of validity, with tightness verified against the linear program in low dimensions. Lemma~\ref{lem:sharp} proves that both lower bounds are attained for every number of treatment values and outcome values, and the upper bounds of \citet[Theorems 4 and 5]{li2024probabilities} stay unchanged.

\begin{lemma}\label{lem:sharp}
For compatible experimental and observational margins,
\begin{gather*}
P(Y_{x_j} = y_i, Y = y_i) \ge \max\{P(x_j, y_i),\ P(y_{i, x_j}) + P(y_i) - 1 + P(x_j) - P(x_j, y_i)\},\\
P(Y_{x_j} = y_i, Y = y_k) \ge \max\{0,\ P(y_{i, x_j}) - P(x_j, y_i) + P(y_k)\\ - P(x_j, y_k) - 1 + P(x_j)\}, \quad k \ne i,
\end{gather*}
and each lower bound equals the probability under some joint distribution of the potential outcomes and the observed variables with the given margins.
\end{lemma}

\begin{proof}
Split the units by the chosen treatment. Units with $X = x_j$ have $Y = Y_{x_j}$ by consistency, so the units choosing $x_j$ contribute $P(x_j, y_i)$ to the first event and nothing to the second. Among the other units, of total probability $1 - P(x_j)$, the event $\{Y_{x_j} = y_i\}$ has probability $a = P(y_{i, x_j}) - P(x_j, y_i)$, the event $\{Y = y_i\}$ has probability $P(y_i) - P(x_j, y_i)$ and the event $\{Y = y_k\}$ has probability $P(y_k) - P(x_j, y_k)$. The Bonferroni inequality within the units with $X \ne x_j$ gives the second terms of both bounds, which proves validity. For attainment, let $a_p$ be the probability of $\{Y_{x_j} = y_i, X = x_p\}$ for $p \ne j$. The margins constrain the $a_p$ only through $\sum_{p \ne j} a_p = a$ and $0 \le a_p \le P(x_p)$, and the overlap with the outcome event within the units choosing $x_p$ is at least $(a_p - s_p)^+$, with the slack $s_p = P(x_p) - P(x_p, y_i)$ for the first event and $s_p = P(x_p) - P(x_p, y_k)$ for the second. Filling the slacks first gives $\sum_{p \ne j} (a_p - s_p)^+ = (a - \sum_{p \ne j} s_p)^+$, which, added to the contribution of the units choosing $x_j$, equals each lower bound. The potential outcomes under the other treatment values are completed independently within every cell of the observed variables, which keeps every margin.
\end{proof}

The second term of the lower bound of \citet[Theorem 4]{li2024probabilities} is smaller than the second term of the lemma by $P(x_j) - P(x_j, y_i)$, and the per-value terms of \citet[Theorem 5]{li2024probabilities} each use the whole set of units with $X \ne x_j$ in place of the units choosing $x_p$. In the binary case with $P(y_x) = 0.9$, $P(x) = 0.5$, $P(x, y) = 0.45$ and $P(y) = 0.6$ the lower bound of \citet[Theorem 4]{li2024probabilities} on $P(Y_x = y, Y = y)$ is $0.50$ and the attained lower bound is $0.55$. The tests of the software compare both lower bounds of the lemma with the exact linear program over the joint distribution of the potential outcomes and the observed variables on random compatible margins.

\msbacksection{CRediT authorship contribution statement}

Marek Deja: Conceptualization, Methodology, Software, Formal analysis, Investigation, Data curation, Writing -- original draft, Writing -- review \& editing.

\msbacksection{Declaration of competing interest}

The author declares no known competing financial interests or personal relationships that could have appeared to influence the work reported in the paper.

\msbacksection{Funding}

The research did not receive any specific grant from funding agencies in the public, commercial, or not-for-profit sectors.

\msbacksection{Data availability}

The protocol, the simulation scripts, the result tables and the software implementing the method (ibstrata 0.2.0, Python, GPL-3.0-or-later) are archived at Zenodo \citep{deja2026bottleneck}, \url{https://doi.org/10.5281/zenodo.24360369}.

\bibliographystyle{elsarticle-num-names}
{\sloppy\bibliography{references}}
\end{document}